\documentclass[10pt,twocolumn,letterpaper]{article}
\usepackage[T1]{fontenc}
\usepackage[pagenumbers]{cvpr} 

\definecolor{cvprblue}{rgb}{0.21,0.49,0.74}
\usepackage[pagebackref,breaklinks,colorlinks,allcolors=cvprblue]{hyperref}
\definecolor{lightblue}{rgb}{0.68,0.85,0.90}

\usepackage{url}

\usepackage[dvipsnames]{xcolor}

\usepackage{amsmath}
\usepackage{amssymb}

\usepackage{booktabs}
\usepackage{multirow}
\usepackage{colortbl}

\usepackage{algorithm} 
\usepackage{algorithmic} 

\usepackage{pifont}
\usepackage{marvosym}
\usepackage{wasysym}

\usepackage{color}

\usepackage{graphicx}
\usepackage{subcaption}
\usepackage{makecell}
\usepackage{cuted}
\usepackage{microtype}
\usepackage{wasysym}
\usepackage{bm}
\usepackage{afterpage} 

\definecolor{green}{HTML}{00B050}
\definecolor{lime}{HTML}{A6CE39}

\title{OmniFashion: Towards Generalist Fashion Intelligence \\ via Multi-Task Vision-Language Learning}

\author{Zhengwei~Yang${^1}$, Andi~Long${^1}$, Hao~Li${^1}$, Zechao~Hu${^1}$, Kui~Jiang${^2}$, Zheng~Wang${^1}$ \\
${^1}$National Engineering
Research Center for Multimedia Software, \\ 
Institute of Artificial Intelligence, School of Computer Science, Wuhan University \\
${^2}$Harbin Institute of Technology \\
{\tt\small yzw\_aim@whu.edu.cn; wangzwhu@whu.edu.cn}
}

\begin{document}
\maketitle
\thispagestyle{fancy}

\begin{abstract}
Fashion intelligence spans multiple tasks, \textit{i.e.,} retrieval, recommendation, recognition, and dialogue, yet remains hindered by fragmented supervision and incomplete fashion annotations.
These limitations jointly restrict the formation of consistent visual–semantic structures, preventing recent vision-language models (VLMs) from serving as a generalist fashion brain that unifies understanding and reasoning across tasks.
Therefore, we construct \textbf{FashionX}, a million-scale dataset that exhaustively annotates visible fashion items within an outfit and organizes attributes from global to part-level.
Built upon this foundation, we propose \textbf{OmniFashion}, a unified vision-language framework that bridges diverse fashion tasks under a unified fashion dialogue paradigm, enabling both multi-task reasoning and interactive dialogue.
Experiments on multi-subtasks and retrieval benchmarks show that OmniFashion achieves strong task-level accuracy and cross-task generalization, highlighting its offering of a scalable path toward universal, dialogue-oriented fashion intelligence.

\end{abstract}

\section{Introduction}
\label{sec:intro}

Fashion plays an increasingly role in digital life, shaping personal identity, influencing consumer behavior, and driving engagement across social and e-commerce platforms~\cite{mohammadi2021smart}. As visual content dominates how fashion is experienced and shared, there is a growing need for intelligent systems that can not only perceive garments but also understand their attributes, recommend relevant styles, retrieve similar items, and answer complex queries interactively~\cite{liao2018knowledge, chen2023multimodal}. These capabilities lie at the intersection of computer vision, natural language processing, and multimodal reasoning, forming the foundation of so-called fashion intelligence~\cite{liu2023toward}.

\begin{figure}
    \vspace{-3mm}
    \centering
    \includegraphics[width=0.999\linewidth]{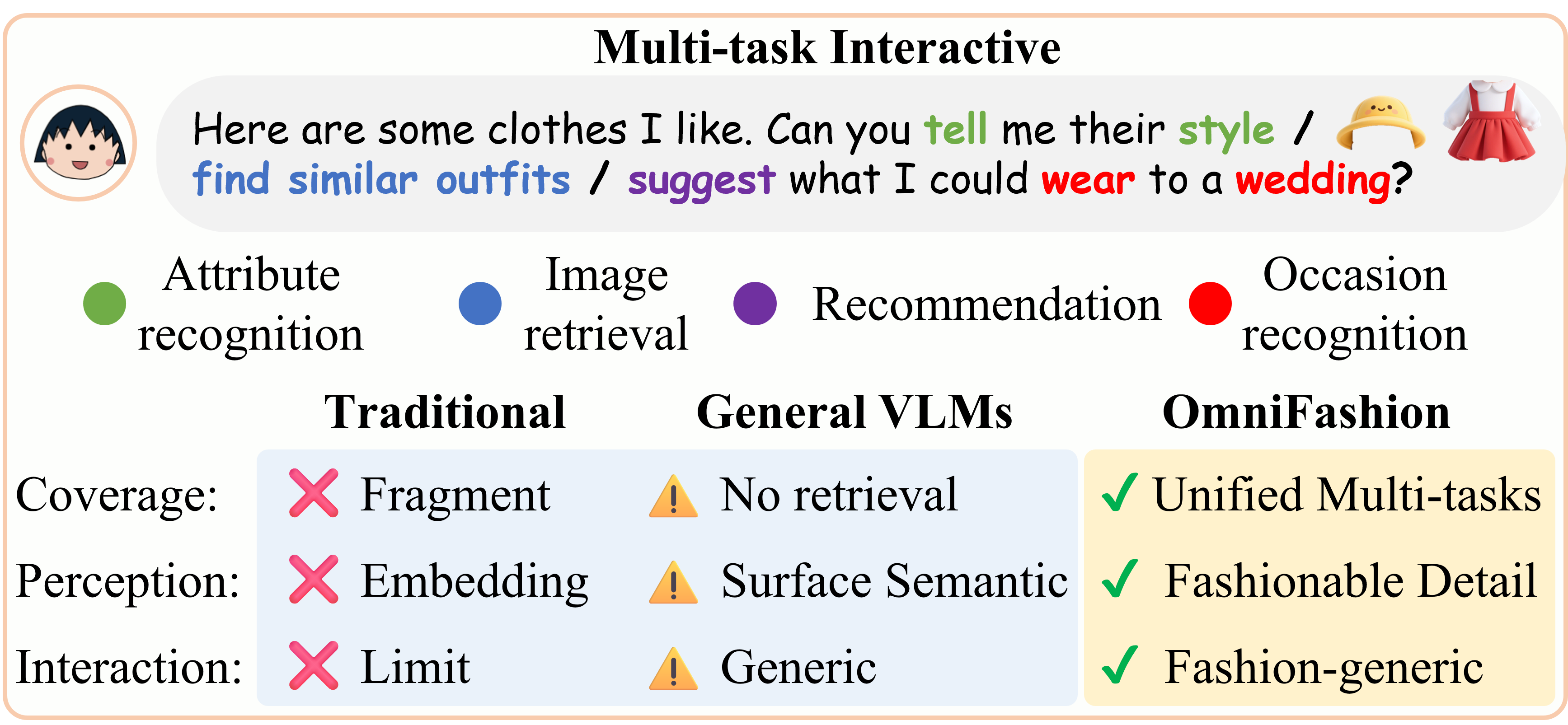}
    \vspace{-5mm}
    \caption{Users interacting with fashion systems require various abilities. Traditional systems address these tasks separately, while general-purpose VLMs offer generic yet shallow responses. OmniFashion unifies multi-task learning with fashion-aware perception and interactive reasoning.}
    \label{fig:motivation}
    \vspace{-3mm}
\end{figure}

Despite growing interest, most existing systems address fashion tasks in fragments. Models trained for retrieval~\cite{goenka2022fashionvlp, han2023fame, chen2024fashionern} struggle to support reasoning or recommendation; A captioning system~\cite{zhao2024unifashion} rarely generalizes to human interaction or contextual understanding. 
As illustrated in Fig.~\ref{fig:motivation}, user needs naturally span multiple objectives, including garment recognition, similarity searching, outfit suggestion, and occasion judgment. 
Traditional methods~\cite{chia2022contrastive, cartella2023openfashionclip, song2024fashiongpt} offer fragmented coverage with limited perception and interaction, and recent attempts to combine related tasks~\cite{han2023fame, pang2025fashionm3, zhao2024unifashion} improve scope but remain loosely coupled.
Meanwhile, general-purpose VLMs~\cite{bai2025qwen2} exhibit broader perception but mostly surface-level understanding, producing generic responses with weak fashion grounding.
This fragmented landscape fails to capture how users naturally interact with fashion content, where diverse goals, from recognizing specific garments to receiving coherent outfit recommendations, coexist and evolve within real-world scenarios~\cite{deldjoo2023review}


\begin{figure*}
\vspace{-3mm}
    \centering
    \includegraphics[width=0.99\textwidth]{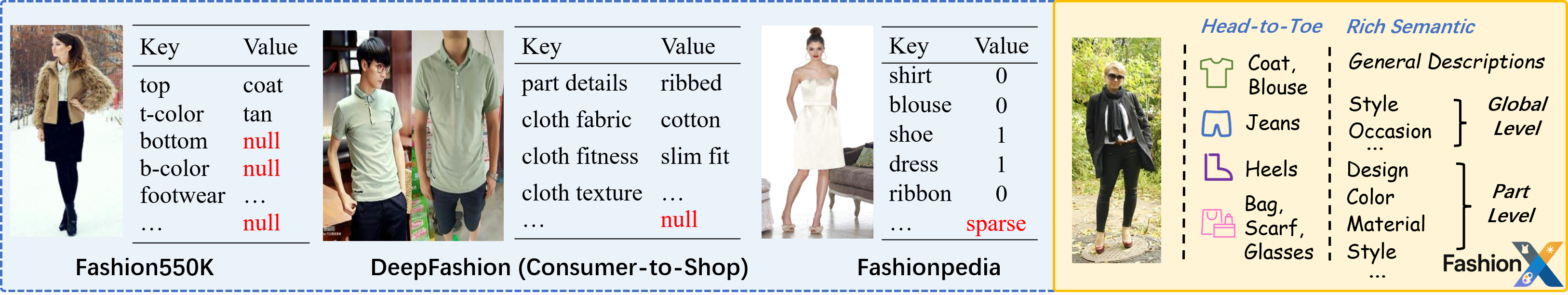}
    \vspace{-3mm}
    \caption{ \textbf{Comparison between existing fashion datasets and FashionX}. Prior datasets show incomplete and inconsistent annotations (marked in red), while FashionX offers a unified head-to-toe coverage with hierarchical annotation structure spanning description, global- and part-level semantics.}
    \label{fig:datasets_cpmpare}
    \vspace{-4mm}
\end{figure*}

This fragmentation originates not only from isolated modeling objectives but also from the data itself. Existing fashion datasets, while valuable, remain incomplete and inconsistent, making unified fashion understanding particularly difficult. In practice, garments rarely appear in isolation: unless in clean product images, clothing is usually worn on models and paired with other items. As illustrated in Fig.~\ref{fig:datasets_cpmpare}, popular datasets such as DeepFashion~\cite{liu2016deepfashion, ge2019deepfashion2}, Fashion550K~\cite{InoueICCVW2017}, Fashionpedia~
\cite{jia2020fashionpedia} exhibits incomplete annotations. 
For instance, an image labeled for a top often also contains bottoms or accessories, yet only the designated item has annotations.
This selective supervision confuses models about which regions correspond to the annotated class and weakens their perception of complete outfits. Moreover, annotation formats vary widely, from categorical tags to binary labeling, hindering coherent cross-task learning. 
To overcome these limitations, we construct \textbf{FashionX}, a million-scale fashion dataset, where annotations are reformulated into consistent image-text pairs that provide head-to-toe annotations with a hierarchical structure, covering from description and fine-grained attributes. This design supplies comprehensive and consistent supervision across diverse objectives, forming a foundation for scalable and unified fashion understanding.

Despite structured data, achieving unified and interactive fashion understanding remains challenging, as it demands coherent integration between global perception and task-specific reasoning. To this end, we present \textbf{OmniFashion}, a comprehensive generalist model for fashion intelligence built upon vision–language foundations. It unifies diverse fashion tasks by reformulating their training objectives into a dialogue-based question–answer paradigm, enabling multi-task collaborative optimization within a single framework. 
Inspired by how humans gradually acquire fashion knowledge, OmniFashion first develops a broad awareness of garments through repeated exposure to varied outfits and descriptions. Then it progressively refines this understanding through focused, goal-oriented multi-task learning, including recommendation, retrieval, attribute recognition, question answering, and reasoning.
This progression allows OmniFashion to develop a context-aware, fine-grained, and generalist understanding of fashion beyond task-specific boundaries.

To realize this progression, OmniFashion builds on the structured supervision of FashionX to construct a comprehensive multi-task setup for unified training.
Fashion intelligence is organized into five major categories, style understanding, scene reasoning, retrieval, attribute recognition, and auxiliary yes/no queries, further divided into thirteen fine-grained subtasks across single- and multi-image, global, and local perspectives.
To reflect how users interact with fashion content, we generate dynamic question–answer variations for each task, capturing diverse querying intents. We further introduce comparative query formats for multi-image retrieval, enabling fine-grained and choice-oriented reasoning that mirrors how users often compare multiple items when making fashion decisions.
These designs enable OmniFashion to learn coherent and consistent representations, supporting generalization across diverse forms of fashion understanding tasks.




Our main contributions are summarized as follows:
\begin{itemize}
    \item We construct FashionX, a million-scale dataset that resolves incomplete and inconsistent annotations in prior resources by providing head-to-toe, hierarchically structured fashion supervision.

    \item We develop OmniFashion, a unified VLM-based multi-task learning framework that achieves context-aware, fine-grained, and generalist fashion understanding through a unified dialogue-based QA training paradigm.
    
    \item Extensive quantitative and qualitative evaluations show that OmniFashion achieves state-of-the-art performance across multiple subtasks and retrieval benchmarks, while demonstrating strong fashion-grounded interactive abilities and multi-image reasoning capabilities.
\end{itemize}

\section{Related Work}
\subsection{Fashion Intelligence}
Fashion intelligence spans a spectrum of vision tasks that aim to understand, interpret, and reason about the fashion items from visual data~\cite{liu2016deepfashion,rostamzadeh2018fashion, ge2019deepfashion2, jia2020fashionpedia, wu2021fashion}. 
A core subtask is fine-grained attribute recognition~\cite{ak2018learning, yang2019interpretable,bhattacharya2022datrnet}, which identifies detailed properties~\cite{ak2018shirt} such as garment type, color, material, and pattern~\cite{wang2018attentive,liu2016fashion}. 
Large-scale annotated datasets have played a key role in advancing supervised methods for structured fashion understanding.
Another major direction is fashion retrieval~\cite{goenka2022fashionvlp,chopra2019powering,xiao2024boosting} and matching~\cite{bai2024beyond,ding2023modeling}, which involves locating visually or semantically related items across different domains, such as consumer-to-shop~\cite{morelli2021fashionsearch} or in-shop~\cite{hadi2015buy}. Early works rely on metric learning~\cite{zhao2019weakly}, part-based alignment~\cite{zhang2022armani}, or attention mechanisms~\cite{tian2023fashion} to handle pose variation and fine-grained differences~\cite{han2022fashionvil}, while recent studies extend retrieval to outfit-level composition~\cite{jang2024lost}, compatibility estimation~\cite{lin2020fashion} and cross-image composition~\cite{tian2025ccin}.
Beyond perception, fashion recommendation~\cite{lin2020outfitnet, sarkar2022outfittransformer, deldjoo2023review} and reasoning~\cite{song2024fashiongpt} have been explored to support personalized outfit analysis, contextual suggestion, and fashion-oriented dialogue~\cite{wang2023fashionvqa, chen2023fashion, song2024fashiongpt}. 
Even with shared semantic foundations, fashion tasks remain fragmented due to task-specific modeling and annotation gaps in available datasets.
These together hinder broader generalization and prevent a unified fashion intelligence from emerging.

\subsection{VLMs for Fashion}
The success of VLMs in general-purpose multimodal understanding has inspired their adoption in the fashion domain. Early efforts adapted pretrained models such as Bert~\cite{zhuge2021kaleido, guan2022personalized, gao2020fashionbert}, CLIP~\cite{radford2021learning, qi2024easy, tian2024expressiveness, han2023fame}, and Diffusion~\cite{ho2020denoising, sun2023sgdiff, baldrati2023multimodal, yan2023fashiondiff}, for captioning, retrieval, VQA and related tasks.
While these models offer strong multimodal priors, downstream adaptation typically depends on task-specific tuning strategies or specialized architectural heads, which restrict scalability across diverse objectives.
These limitations have led to a growing interest in fashion-oriented VLMs that encode fashion semantics more directly. Some incorporate attribute alignment~\cite{chia2022contrastive, han2022fashionvil, cartella2023openfashionclip, kanade2024fashion, dong2025open}, hierarchical category modeling~\cite{ngom2024enhancing}, or multimodal catalog data~\cite{chen2023fashion, song2024fashiongpt, pang2025fashionm3} to expand fashion domain grounding.
Others incorporate auxiliary signals such as garment codes~\cite{bian2025chatgarment}, 3D geometry~\cite{feng2024chatpose}, or user intents~\cite{qiu2022pre} to enable controllable generation and try-on applications~\cite{song2025image, shi2025generative, xu2025personalized}.
Although these fashion VLMs extend beyond single-task settings, they remain largely task-oriented, with objectives handled through separate training setups. We instead adopt a unified formulation that enables diverse fashion tasks to be optimized within a single model.



\section{Methods}

OmniFashion builds on FashionX to support unified vision–language learning.
We first show the FashionX annotation pipeline in Sec.~\ref{sec:3.1}, which produces large-scale annotation with hierarchical semantics. 
Sec.~\ref{sec:3.2} then introduces the unified task formulation and progressive training strategy, detailing how diverse fashion tasks are cast into a uniform paradigm.
The inference setup is outlined in Sec.~\ref{sec:3.3}. Finally, Sec.~\ref{sec:3.4} provides statistical analyses of FashionX, summarizing the task composition, outlining how it supports both alignment and unified multi-task learning.


\begin{figure}
    \centering
    \includegraphics[width=0.99\linewidth]{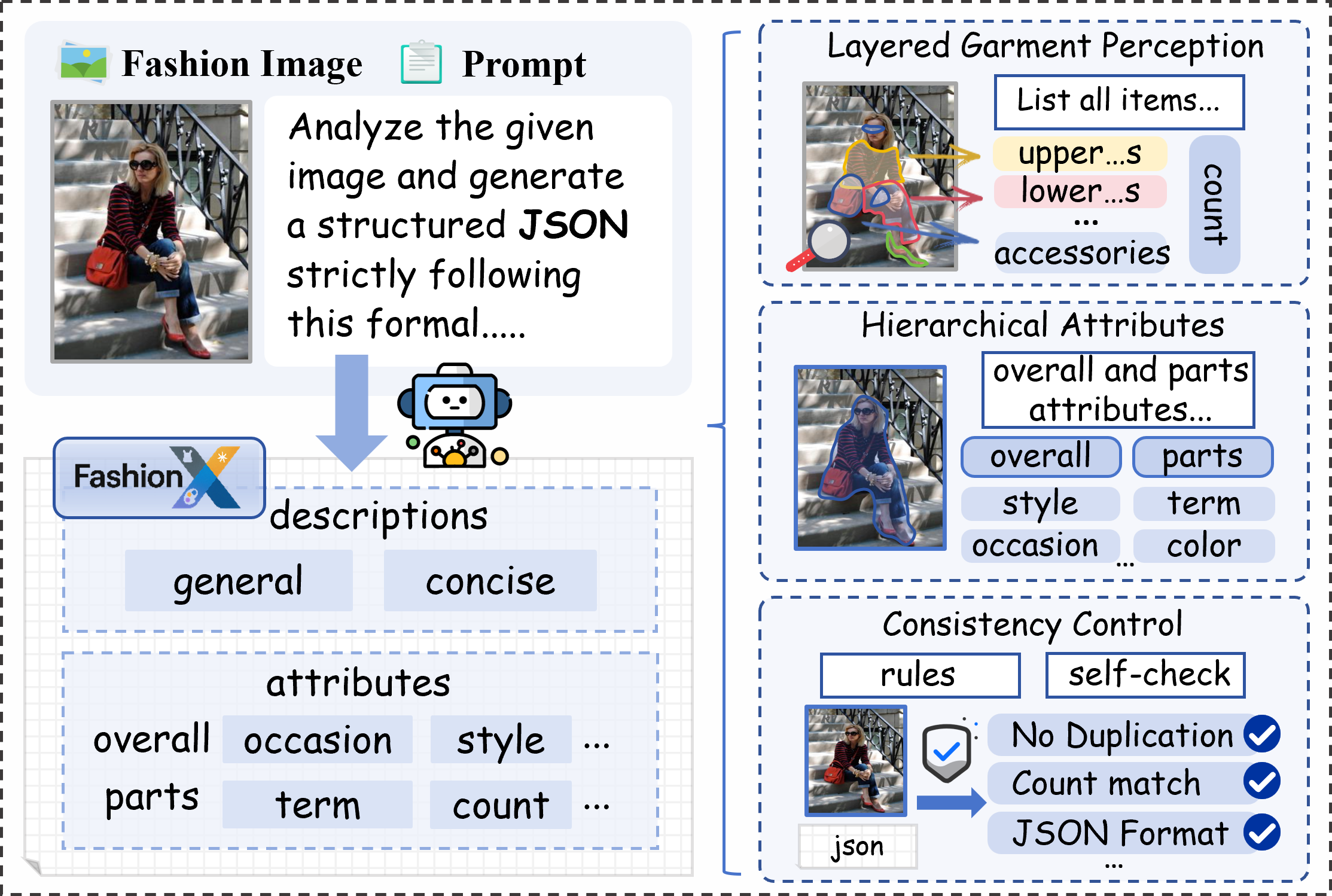}
    \caption{Overview of the FashionX annotation pipeline.}
    \vspace{-3mm}
    \label{fig:annotation}
    \vspace{-3mm}
\end{figure}

\subsection{Automated Annotation Pipeline of FashionX}
\label{sec:3.1}
To address the incomplete, inconsistent annotations in existing fashion datasets, we develop an automated VLM-based annotation pipeline to construct FashionX, as illustrated in Fig.~\ref{fig:annotation}.
Unlike generic captioning pipelines, the system is designed for the compositional and layered nature of fashion images, where multiple garments appear simultaneously and interact through various attributes.
Given a raw fashion image, a structured prompt guides the VLM to generate JSON-formatted annotations composed of two complementary components: descriptions (general and concise \footnote{The general part focuses on the target garments, whereas the concise part includes broader contextual cues such as environment or activity.}) capturing global outfit semantics, and hierarchical attributes covering overall characteristics (e.g., style, occasion) and part-level details (e.g., term, color, material, pattern). This unified structure provides complete head-to-toe coverage and preserves the fine-grained garment relationships needed for downstream multi-task learning.

Although the pipeline follows a standard prompted generation template, three key adaptations ensure coverage and consistency:
\textbf{Layered garment enumeration:} The prompt explicitly requires listing and counting all visible garments, ensuring that co-occurring items and layered outfits are fully captured.
\textbf{Hierarchical attribute decomposition:} Attributes are organized into overall and part-level groups to represent abstract outfit semantics and concrete garment details in a coherent structure.
\textbf{Consistency control:} A built-in self-check verifies category non-duplication, alignment between counted and listed items, and strict JSON formatting.
These components collectively enable FashionX to yield over one million unified and self-validated image–text pairs, providing a reliable foundation for generalist fashion.

\subsection{Training of OmniFashion}
\label{sec:3.2}

\begin{figure}[H]
\vspace{-5mm}
    \centering
    \includegraphics[width=0.9\linewidth]{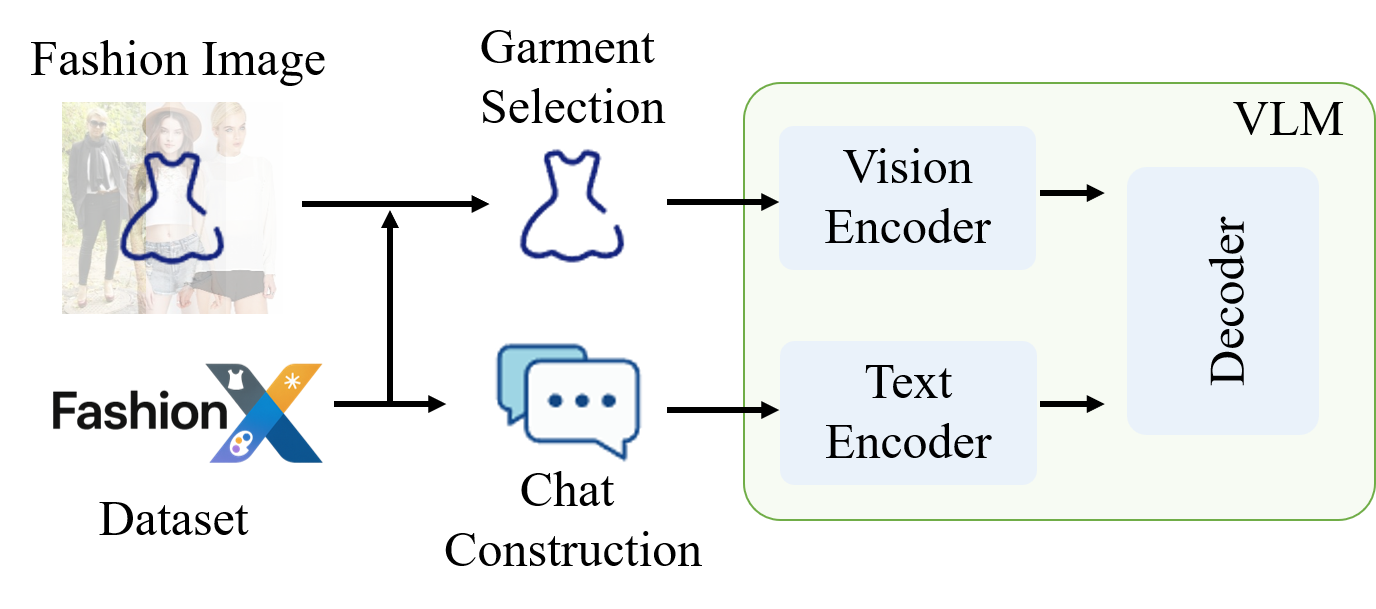}
    \vspace{-3mm}
    \caption{\textbf{Pipeline of OmniFashion.} OmniFashion builds on FashionX datasets that first construct data with garment and corresponding description/attribute. The VLM output will be penalized by the constructed dialogue as an answer.}
    \label{fig:structure}
    \vspace{-3mm}
\end{figure}

\begin{figure*}
    \vspace{-3mm}
    \centering
   \includegraphics[width=0.99\textwidth]{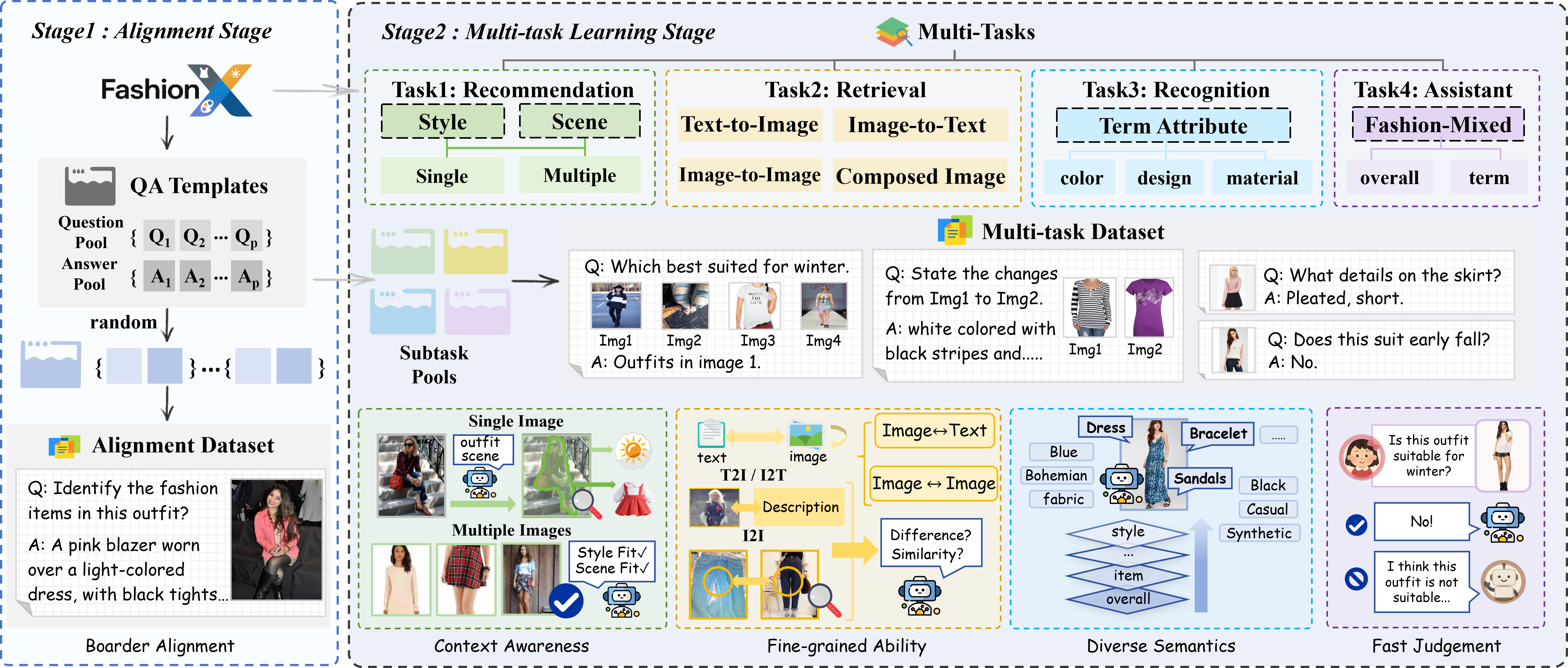}
    \vspace{-2mm}
    \caption{Illustration of the progressive learning task design in OmniFashion. Different color stands for different tasks.}
    \label{fig:2stage_training}
    \vspace{-3mm}
\end{figure*}

Fig.~\ref{fig:structure} illustrates the training pipeline of OmniFashion. Instead of relying on task-specific heads, all fashion objectives are converted into a unified dialogue QA format, allowing the model to learn retrieval, recognition, and reasoning within the same conversational interface. Each training sample consists of various fashion images and a query-answer pair derived from FashionX annotations.

\subsubsection{Problem Formulation}
Formally, given an image input $X_v$ and a text query $T_q$ raised in dialogue, the model predicts a textual response $Y$ as:
\begin{equation}
    Y = f_{\theta}(X_v, T_q)
\end{equation}
where ${\theta}$ denotes the tunable parameter of vision-language backbone such as Qwen2.5-VL~\cite{bai2025qwen2}. 
All tasks share the same autoregressive generation loss:
\begin{equation}
\mathcal{L} = -\frac{1}{N} \sum_{i=1}^{N} \log P(Y_i | X_{v,i}, T_{q,i}).
\end{equation}

Based on this unified QA formulation, OmniFashion is trained by first focusing on fashion-oriented visual–semantic grounding, and then extending to diverse downstream tasks through multi-task QA supervision.

\subsubsection{Visual-semantic Alignment Stage}
As shown in the right of Fig.~\ref{fig:2stage_training}. This stage aims to establish fashion-aware visual–semantic grounding through unified dialogue QA fine-tuning. 
While pretrained VLMs offer strong general representations, their understanding of fashion-specific concepts, such as garment layering and styling cues, remains implicit.
To refine this grounding, given each fashion image $X_v$, a question template $T_q$ is sampled from a question pool $\mathcal{Q}$, and an answer template is selected from a corresponding answer pool $A$ as:
\begin{equation}
     T_q \sim \mathcal{Q}, \qquad Y_a \sim \mathcal{A},
\end{equation}
where the instantiated answer $Y_a$ is filled by the general description annotations from FashionX to maintain a clear focus on fashion semantics.

This template-based construction yields a diverse set of QA forms for the same supervisory signal, encouraging the model to internalize fashion semantics as dialogue-relevant concepts rather than static labels.
The resulting alignment dataset provides broad fashion grounding and prepares the model for unified multi-task dialogue learning.

\subsubsection{Unified Multi-Task Learning}

The second phase applies the unified formulation to a broader set of fashion objectives. 
Each task is instantiated through its own question and answer template pools, from which QA pairs are sampled in a manner consistent with the alignment stage; the specific template designs are omitted for brevity.
The following sections describe how different fashion tasks are constructed under this unified QA paradigm.

\paragraph{Fashion Recommendation}
The recommendation tasks assess the model’s ability to reason about style compatibility and scene appropriateness.
As shown in the green region of Fig.~\ref{fig:2stage_training}, each objective is expressed in both single-image and multi-image forms, resulting in four complementary QA settings that reflect different ways users evaluate outfits.
For style recommendation, FashionX provides signals at two levels: overall style describing the outfit as a whole, and part style characterizing individual garments, which allows queries to reference either holistic appearance or specific items.
Scene-based queries follow the same single-image and multi-image structures to determine whether an outfit matches a described scenario.

Formally, in single-image cases, the model answers a style- or scene-related query $T_q^{\text{rec}}$ given a single outfit $X_v$:
\begin{equation}
    \hat{Y}_a = f_\theta(X_v,\, T_q^{\text{rec}}),
\end{equation}
where $\hat{Y}_a$ is the predicted output, we reuse this term in the following, since all settings follow the same generative formulation and produce natural-language responses.

For multi-image cases, the query is posed over a set of candidate images simultaneously, formulated as:
\begin{equation}
    \hat{Y}_a  = f_\theta(\{X_v^1, X_v^2, \ldots, X_v^k\},\, T_q^{\text{rec}}),
\end{equation}
where the multi-image design reflects real fashion decision-making, where users frequently compare multiple outfits and ask for preferences.
And together, these formulations move recommendations beyond static classification, enabling the model to perform context-aware aesthetic reasoning that integrates garment-level cues with holistic outfit coherence.


\paragraph{Fashion Retrieval}

As illustrated in the yellow region of Fig.~\ref{fig:2stage_training}, retrieval tasks are reformulated by converting the original query–gallery structure into dialogue pairs that explicitly ask for the correct visual–textual correspondence.
This includes text-to-image, image-to-text, and image-to-image variants, all expressed through natural-language queries that reference possible candidate items depending on the subtasks.
Questions take forms such as identifying which image matches a description, selecting the description aligned with an outfit, or deciding whether two images depict the same garment or stylistic variant. Beyond these direct correspondence queries, we further include a comparative retrieval form in which the model needs to articulate differences between two images. This comparative setting strengthens the model’s sensitivity to subtle fashion cues.
Formally, given a query set $\mathcal{Q}$ and a candidate set $\mathcal{C}$, retrieval is expressed as:
\begin{equation}
\hat{Y}_{a} = f_\theta \left(\{\mathcal{Q},\mathcal{C}\}, T_q^{\text{ret}} \right),
\end{equation}
where $T_q^{\text{ret}}$ specifies the retrieval instruction. 


\paragraph{Fashion Recognition}
As illustrated in the cyan region of Fig.~\ref{fig:2stage_training}, recognition tasks focus on part-level garment attributes, guiding the model to describe properties such as color, design details, and material. Each subtask is framed as a natural-language query referring to a specific garment part, prompting the model to express the relevant attribute in descriptive form. 
This encourages the model to connect visual appearance with semantically distinct attribute categories, enabling it to articulate how a garment looks, how it is constructed, and what it is made of.
Formally, given an attribute-oriented query $T_q^{\text{attr}}$, recognition is written as:
\vspace{-2mm}
\begin{equation}
\hat{Y}_a = f_\theta(X_v, T_q^{\text{attr}}).
\vspace{-3mm}
\end{equation}

\paragraph{Quick Assistance}

As shown in the purple region of Fig.~\ref{fig:2stage_training}, the quick-assistance tasks focus on deterministic fashion judgments, where the model must decide whether an outfit satisfies a specific semantic condition, such as style, occasion, season, or a given attribute.
Unlike descriptive queries, these questions require concise binary answers drawn from the template pool, prompting the model to resolve clear semantic boundaries rather than generate free-form descriptions. Formally, given a condition-driven query $T_q^{\text{aux}}$, the task is expressed as:
\begin{equation}
\hat{Y} = f_\theta(X_v, T_q^{\text{aux}}).
\end{equation}

\subsection{Inference of OmniFashion}
\label{sec:3.3}
At inference, OmniFashion uses the same QA-based formulation as in training. For every subtask, including recommendation, recognition, reasoning, and retrieval, we construct a held-out test split in the same format as the training data. To ensure stability and reproducibility, we employ fixed prompt templates across tasks. This setup evaluates generalization without any task-specific adapters or post-hoc tuning.

For conventional retrieval benchmarks, directly querying a VLM against the full gallery is impractical, due to the high computational cost of exhaustive image–text evaluation.
Recent success on model rewarding~\cite{chiang2024chatbot, sun2024rethinking} inspired us to develop a Bradley–Terry~\cite{bradley1952rank} matching strategy. 
The model performs tournament-style pairwise comparisons within small candidate pools, and the resulting pairwise preferences are aggregated to derive a global ranking.  
This approach preserves the QA inference format while enabling direct evaluation against conventional retrieval systems.

\subsection{Data Analysis}
\label{sec:3.4}

\begin{table}[h]
    \centering
    \scriptsize
    \vspace{-3mm}
    \caption{Statistical Detail of FashionX. Abbreviations: One-piece (O), top (T), bottom (B), footwear (F), accessory (A).}
    \vspace{-3mm}
    \setlength{\tabcolsep}{2pt}
    \begin{tabular}{l|ccccccc}
    \toprule
     & O & T & B & F & A & Outfits \\
    \midrule
    Num.     &261,389 & 952,229    & 729,562 & 593,448 & 797,970 & 1,032,710 \\
    Attri. Num.  & 2,656,329 & 7,855,390 & 5,099,191 & 3,964,328 & 4,533,838 & 6,135,520 \\
    \midrule
    Total Items &  \multicolumn{6}{c}{3,334,598} \\
    Total Attri. &  \multicolumn{6}{c}{24,109,076} \\
    \bottomrule
    \end{tabular}
    \label{tab:dataset_detail}
    \vspace{-3mm}
\end{table}

\textbf{FashionX} comprises 1,027,710 fashion-related outfits collected and annotated through the automated pipeline described in Sec.~3.1. The data originate from multiple fashion benchmarks~\cite{liu2016deepfashion, jia2020fashionpedia, InoueICCVW2017} and cover a wide range of clothing types, styles, and contextual scenes.
All annotations are generated by GPT-4.1~\cite{achiam2023gpt} under a unified JSON schema, producing both global- and part-level labels across one-piece, tops, bottoms, footwear, and accessories.
The dataset is split 90\%/10\% into training and testing with no outfit overlap. As summarized in Tab.~\ref{tab:dataset_detail}, FashionX offers broad semantic coverage, containing millions of garment items and over twenty million attribute annotations across major clothing categories, forming a comprehensive foundation for downstream alignment and multi-task construction.


\begin{table*}[t]
	\centering
    \small
	\caption{Comparison with the SOTA methods across eight fashion dialogue subtasks, including overall style, part style, occasion, color recognition, and four retrieval dialogue settings (I2T, T2I, I2I, CIR). The table reports both A@1 and Acc metrics. OmniFashion shows consistently strong performance across all subtasks and achieves the highest mean accuracy among the open- and close-source models.}
    \vspace{-2mm}
	\setlength{\tabcolsep}{5pt}
	\begin{tabular}{lc|cccccccc|cccc|c}
		\toprule
		\multirow{2}[2]{*}{Methods} & 
		\multirow{2}[2]{*}{Size} 
		& \multicolumn{2}{c}{Overall Style} & \multicolumn{2}{c}{Part Style} 
        & \multicolumn{2}{c}{Occasion} 
        & \multicolumn{2}{c}{Color} 
        & {I2T} 
        & {T2I} 
        & {I2I} 
        & {CIR}
        & \multirow{2}[2]{*}{\makecell[c]{Mean \\ Acc}}\\
		\cmidrule(lr){3-4} \cmidrule(lr){5-6} \cmidrule(lr){7-8} \cmidrule(lr){9-10}  \cmidrule(lr){11-14}  
		& & A@1 & Acc & A@1 & Acc & A@1 & Acc & A@1 & Acc & Acc & Acc  & Acc & Acc  \\
        \hline
        \rowcolor{gray!15} \textbf{\textit{Close-sourced VLMs}} & & & & & & & & & & & & & & \\

      \rowcolor{gray!15}
        \color{gray!85}Grok-4-Fast & \color{gray!85}- & \color{gray!85}38.0 & \color{gray!85}66.6 & \color{gray!85}11.6 & \color{gray!85}20.3 & \color{gray!85}87.8 & \color{gray!85}91.5 & \color{gray!85}67.1 & \color{gray!85}69.3 & \color{gray!85}90.4 & \color{gray!85}20.6 & \color{gray!85}10.7 & \color{gray!85}15.4 & \color{gray!85}48.1\\
        \rowcolor{gray!15}
        \color{gray!85}Gemini-2.5-Flash & - & \color{gray!85}75.3 & \color{gray!85}89.8 & \color{gray!85}10.7 & \color{gray!85}10.9 & \color{gray!85}92.0 & \color{gray!85}94.9 & \color{gray!85}78.8 & \color{gray!85}81.2 & \color{gray!85}99.7 & \color{gray!85}99.3 & \color{gray!85}60.9 & \color{gray!85}19.3 & \color{gray!85}69.5 \\
        \rowcolor{gray!15}
        \color{gray!85}Claude-4.5-Sonnet & \color{gray!85}- & \color{gray!85}74.6 & \color{gray!85}91.2 & \color{gray!85}32.4 & \color{gray!85}40.5 & \color{gray!85}89.7 & \color{gray!85}99.2 & \color{gray!85}67.9 & \color{gray!85}70.6 & \color{gray!85}98.5  & \color{gray!85}95.8 & \color{gray!85}58.6 & \color{gray!85}23.2 & \color{gray!85}72.2 \\
        \hline
        \textbf{\textit{Open-sourced VLMs}} & & & & & & & &  & & & & &  \\
        MiniCPM-V 2.6~\cite{yao2024minicpm} & 8b & 58.3 & 59.5 & 21.7 & 24.2 & 34.6 & 60.7 & 29.4 & 40.2 & 59.5 & 98.1 & 37.3 & 53.9 & 54.2 \\
        LLaVA-OneVision-1.5~\cite{LLaVA-OneVision-1.5} & 8b & 85.5 & 92.7 & 20.1 & 41.7& 86.4 & 94.8 & 72.7 & 84.2 & 99.2 & 25.1 & 11.3 & 46.1 & 61.9 \\
        Qwen2.5-VL~\cite{bai2025qwen2} & 7b & 20.8 & 68.8 & 19.1 & 48.4 & 90.4 & 90.6 & 71.2 & 71.3 & \textbf{99.5} & 97.4 & 39.3 & 37.5 & 69.1 \\
        Phi-4-multimodal~\cite{abouelenin2025phi} & 5b & 50.4 & 66.9 & 15.1 & 17.9 & 86.7 & 98.1 & 18.9 & 20.5 & 96.7 & 77.9 & 46.2 & 48.2 & 59.0\\
        LLaVA-OneVision-1.5~\cite{LLaVA-OneVision-1.5} & 4b & 83.2 & 91.7 & 12.4 & 24.2 & 84.1 & 94.8 & 70.0 & 80.7 &  98.7 & 20.7 & 11.1 & 30.0 & 56.5 \\
        Qwen2.5-VL~\cite{bai2025qwen2} & 3b & 48.9 & 80.6 & 21.1 & 42.5 & 76.2 & 76.2 & 68.7 & 68.7 & 81.4 & 28.5 & 9.8 & 31.9 & 52.4 \\
        \hline
        \rowcolor{lightblue!30} OmniFashion (ours) & 3b  & \textbf{86.0} & \textbf{93.4} & \textbf{66.2} & \textbf{73.5} & \textbf{89.7} & \textbf{97.0} & \textbf{79.1} & \textbf{86.8} & \textbf{99.5} & \textbf{98.7} & \textbf{89.4} & \textbf{75.4} & \textbf{87.8} \\
        \hline
	\end{tabular}
	\label{tab:vqa}
    \vspace{-3mm}
\end{table*}

\textbf{Visual-semantic alignment stage} utilizes the entire 900K training split of FashionX to enhance fashion-specific visual–semantic grounding.
Each sample is reformulated into dialogue QA pairs, resulting in one million training examples, including 100K general dialogue samples from LLaVA-Instruct~\cite{liu2023visual} to preserve conversational fluency.

\textbf{Multi-task learning stage} builds on the same training images but constructs task-specific QA pairs from FashionX, according to the four major and thirteen sub-task, ensuring a balanced composition among them:
\begin{itemize}
    \item Style / Scene: 150 K / 150 K QA pairs respectively, covering overall, part-level, and multi-context formulations.
    \item Retrieval: 400 K QA pairs, devised to 100K for each task, including text-to-image, image-to-text, image-to-image, and composed image retrieval.
    \item Attribute Recognition: 150 K QA pairs, equally divided among color, design, and material attributes.
    \item Assistance: 100 K binary QA pairs, half focusing on global (scene, season) and half on local (color, design) judgments.
    \item Additional Dialogue: 120K general dialogue samples from LLaVA-Instruct~\cite{liu2023visual}
\end{itemize}
In total, it contributes over one million QA instances, complementing the alignment stage and forming a unified, large-scale corpus for multi-task fashion intelligence training.

\section{Experiments}
\subsection{Experimental Setup}
\paragraph{Datasets} All experiments are conducted on the FashionX dataset introduced before. For evaluation, we construct a held-out test set from the FashionX corpus, covering a broad range of fashion subtasks to assess recognition and reasoning capability. Specifically, the test split includes 20K samples for overall scene understanding, 50K for overall style, 50K for part-level style, 50K for part-level color recognition, 100K for image-to-text retrieval, 100K for text-to-image retrieval, 50K for image-to-image retrieval, and 6K for composed image retrieval.
All test samples are formatted in the same QA structure as used in training to ensure consistent evaluation.
For retrieval-related experiments, we additionally adopt the same test splits as the DeepFashion in-shop and consumer-to-shop benchmarks~\cite{liu2016deepfashion}.
These retrieval samples are not included in the FashionX training split, ensuring non-overlap between training and evaluation data.
OmniFashion is implemented on top of the Qwen2.5-VL backbone and fine-tuned using LoRA adaptation for efficient parameter updating. Detailed fine-tuning configurations are provided in the Appendix for reproducibility.

\paragraph{Evaluation Metric}
For dialogue tasks, the model generates up to three candidate answers per question. We report A@1, the percentage whose first answer matches the ground truth, and Acc, the percentage where any generated answer is correct. 
For the CIR task, correctness is determined by CLIP text–text similarity between the prediction and reference, with threshold of 0.75.
For retrieval, we evaluate two settings: (1) the dialogue-based QA, which verifies the model’s learned matching ability through interaction, and (2) the standard retrieval protocol from DeepFashion~\cite{liu2016deepfashion}, reporting Recall@K (R@K) and mean Average Precision (mAP) for direct comparison with conventional retrieval methods.

\subsection{Performance Comparison}

\paragraph{Fashion dialogue-based Tasks} 
Tab.~\ref{tab:vqa} reports comparisons with both open- and closed-source VLMs across reasoning, recognition, and retrieval QA tasks. Several consistent patterns emerge.
A clear pattern emerges when examining the behavior of existing VLMs.
Tasks grounded in broadly shared visual knowledge, such as recognizing basic colors or judging simple scene suitability, are handled relatively well across most models.
When the evaluation shifts toward style-related reasoning, performance often declines, suggesting that general-purpose pretraining provides only limited support for fashion-specific conceptual understanding.

\begin{table}[t]
\vspace{-3mm}
	\centering
    \footnotesize
	\caption{\small Comparison on the InShop retrieval benchmark.}
    \vspace{-2mm}
	\setlength{\tabcolsep}{8pt}
	\begin{tabular}{l|cccc}
		\toprule
		\multirow{2}[2]{*}{Methods} 
        & \multicolumn{4}{c}{InShop}  \\
		\cmidrule(lr){2-5} 
		 & R@1 & R@10 & R@20 & mAP   \\
        \midrule
        Hyper-DINO~\cite{ermolov2022hyperbolic} & 92.6 & 98.4 & 99.0 & -   \\
        Hyper-ViT~\cite{ermolov2022hyperbolic} & 92.7 & 98.4 & 98.9 & -   \\
        IBC~\cite{seidenschwarz2021learning}  & 92.8 & 98.5 & 99.1 & - \\
        DADA~\cite{ren2024towards} & 93.0 & 98.5 & 98.9 & -\\
        MGA~\cite{zhu2023fashion} & 94.3  & 98.8 & 99.1 & 81.1 \\
        \midrule
        OmniFashion(Ours)  & \textbf{95.2} & \textbf{99.2} & \textbf{99.3} & \textbf{82.5} \\
    \bottomrule
	\end{tabular}
	\label{tab:retrieval_inshop}
    \vspace{-2mm}
\end{table}

\begin{table}[t]
	\centering
    \footnotesize
	\caption{Comparison on the Consumer2Shop retrieval benchmark.}
    \vspace{-2mm}
	\setlength{\tabcolsep}{8pt}
	\begin{tabular}{l|cccc}
		\toprule
		\multirow{2}[2]{*}{Methods} 
        & \multicolumn{4}{c}{Consumer2Shop}  \\
		\cmidrule(lr){2-5} 
		& R@1 & R@10 & R@20 & mAP   \\
        \midrule
        Ensemble~\cite{hauri2025virtual} & 23.1 & 48.0 & - & - \\
        RTS~\cite{wieczorek2020strong} & 37.8 & 71.1 & 77.2 & 43.0 \\
        CTL~\cite{wieczorek2021unreasonable} & 37.6 & 71.1 & 77.6 & 43.1 \\
        MGA~\cite{zhu2023fashion}  & 38.4 & 72.3 & 78.3 & 44.2 \\
        Hyper-DINO~\cite{ermolov2022hyperbolic} & 42.5 & 65.9 & - & - \\
        \midrule
        OmniFashion(Ours) & \textbf{43.9} & \textbf{75.9} & \textbf{80.8} & \textbf{48.1} \\
    \bottomrule
	\end{tabular}
	\label{tab:retrieval_c2shop}
    \vspace{-4mm}
\end{table}

Retrieval-oriented QA exposes a second trend.
Most models are more reliable when matching an image to a text description than when retrieving an image from text.
This discrepancy becomes more apparent as the task requires finer discrimination.
Image–image comparison poses particular challenges, since the model must contrast multiple outfits simultaneously and detect subtle changes across garments.
The composed retrieval task further amplifies this difficulty, where many models fail to describe attribute-level differences consistently.
Within this landscape, OmniFashion attains the strongest overall accuracy among open-source VLMs and performs competitively with larger closed-source systems.
The improvements are especially visible in style reasoning and multi-image comparison, indicating that the structured supervision of FashionX equips the model with more stable, context-aware, and fine-grained fashion understanding than generic multimodal training alone.

\paragraph{Fashion Retrieval} 
As shown in Tabs.~\ref{tab:retrieval_inshop} and ~\ref{tab:retrieval_c2shop}, OmniFashion achieves superior retrieval performance on both the Deepfashion-InShop and Deepfashion-Consumer2Shop benchmarks. On the InShop dataset, it obtains the highest scores across all metrics, reaching 95.2\% in R@1, and 82.5\% in mAP, outperforming recent methods such as MGA~\cite{zhu2023fashion} and DADA~\cite{ren2024towards}. This improvement indicates stronger fine-grained matching between clothing items, achieved without any task-specific retrieval head. On the more challenging Consumer2Shop benchmark, OmniFashion also surpasses the second-best approaches~\cite{zhu2023fashion, ermolov2022hyperbolic}, improving R@1 from 42.5\% to 43.9\% and mAP from 44.2\% to 48.1\%, demonstrating better cross-domain generalization and alignment from consumer photos to shop images. Overall, these results verify that the QA-based training of OmniFashion effectively transfers to standard retrieval tasks, yielding discriminative yet semantically consistent fashion representations that outperform traditional retrieval-focused architectures.
\begin{figure}
\vspace{-3mm}
    \centering
    \includegraphics[width=0.99\linewidth]{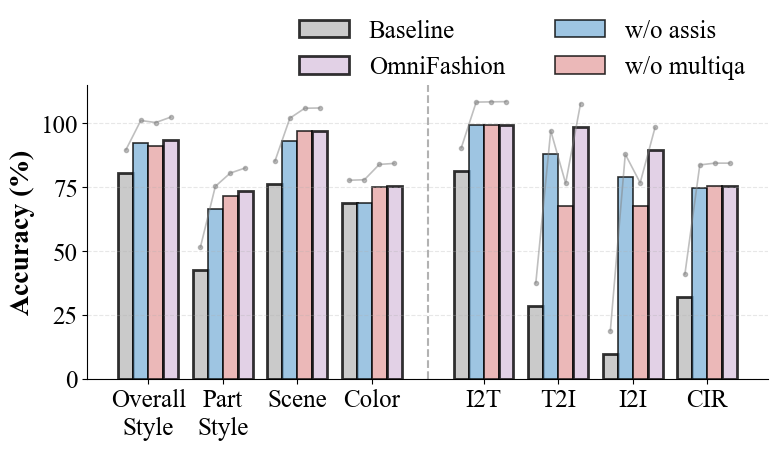}
    \vspace{-3mm}
    \caption{\textbf{Ablation on the task variants.} Training without assistance tasks (blue, w/o assis) or multi-image recommendation tasks (red, w/o multiqa) are shown with accuracy on all subtasks. The gray trend lines upper the bars in each group indicate performance changes relative to the baseline. }
    \label{fig:ablation}
    \vspace{-4mm}
\end{figure}

\begin{figure}
    \centering
    \includegraphics[width=0.99\linewidth]{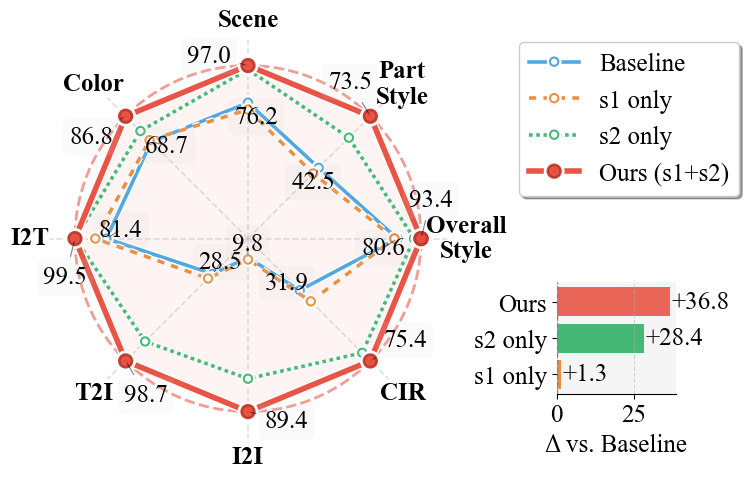}
    \vspace{-4mm}
    \caption{\textbf{Ablation on the two-stage training.} The radar chart compares accuracy across tasks for models trained with s1 only, s2 only, and both stages (ours); the bar chart summarizes overall gains vs. the baseline. Red (solid) indicates the full model, yellow (dashed) and green (dotted) denote s1-only and s2-only variants, respectively, with blue showing the baseline.}
    \label{fig:ablation_radar}
    \vspace{-4mm}
\end{figure}

\begin{figure*}
    \centering
    \includegraphics[width=0.95\textwidth]{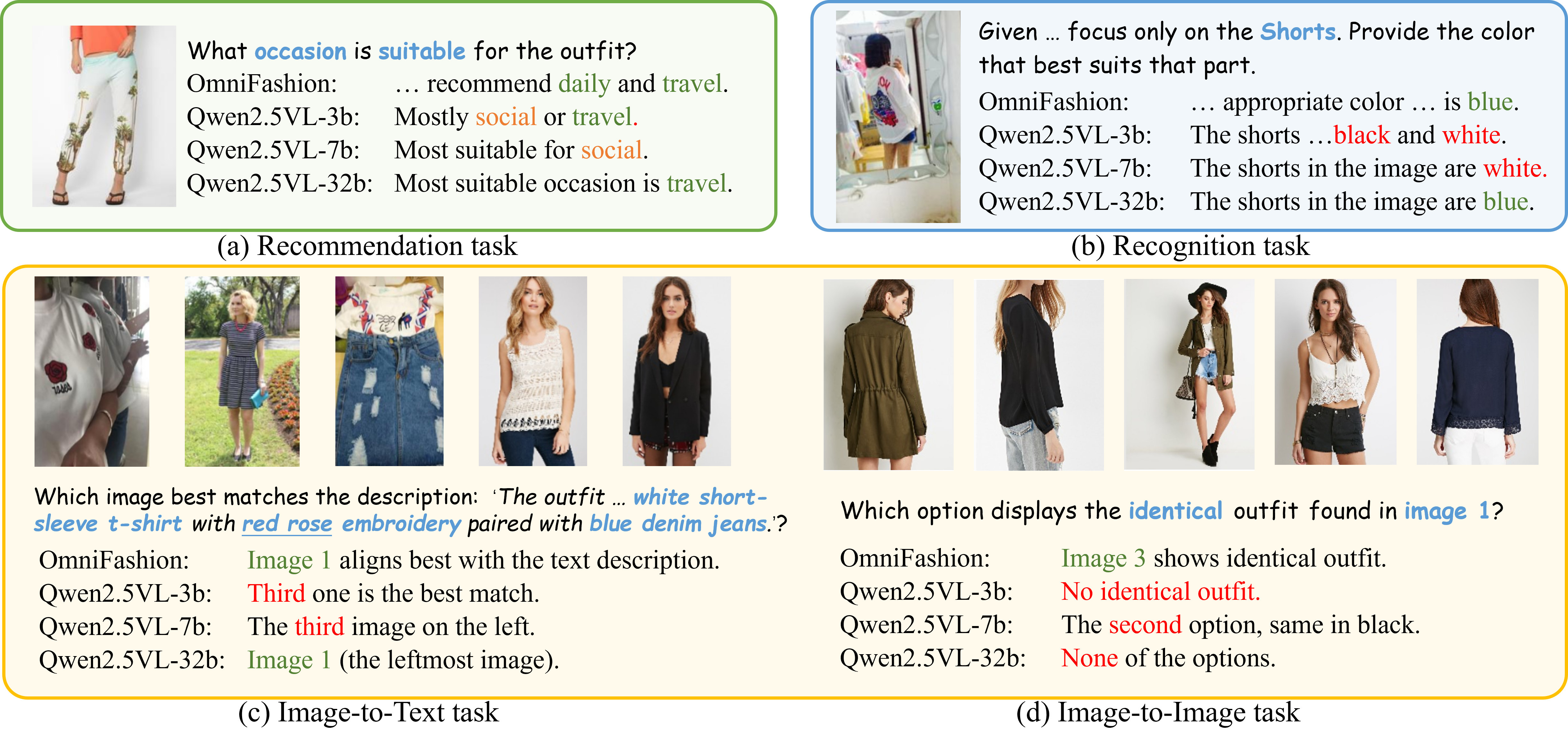}
    \vspace{-2mm}
    \caption{\textbf{Qualitative comparison across subtasks.} Each block shows input queries and model responses. Colored text highlights correct (\textcolor{green}{green}), incorrect (\textcolor{red}{red}), and uncertain (\textcolor{orange}{orange}) predictions, while underlined phrases mark key visual cues. }
    \vspace{-3mm}
    \label{fig:qualitative}
\end{figure*}

\subsection{Ablation Studies}
\paragraph{Effectiveness of Subtasks} Figure~\ref{fig:ablation} shows the ablation analysis of assistance and multi-image recommendation subtasks.
Removing assistance tasks (blue bars) present a clear drop in reasoning and recognition accuracy, especially for overall style and color, showing that the assistance tasks help maintain semantic and attribute consistency. Ignoring multi-image recommendation (red bars) mainly weakens Text-to-Image and Image-to-Image performance, confirming its role in improving comparative reasoning across multiple visual references. As indicated by the gray trend lines, both modules provide complementary benefits: assistance tasks refine fine-grained reasoning, while multi-image recommendation enhances relational understanding in multi-image scenarios.


\paragraph{Effectiveness of Training Stage}
Figure~\ref{fig:ablation_radar} illustrates the ablation results of our two-stage training framework.
Using only Stage 1 (s1) yields marginal gains, mainly on retrieval QA tasks, suggesting that pure image–text alignment provides limited improvement for fashion-specific understanding.
In contrast, Stage 2 (s2) brings significant accuracy gains across nearly all tasks, confirming the benefit of fashion-focused multi-task training; however, it still underperforms the full two-stage framework, especially on Part Style, Text-to-Image, and Image-to-Image retrieval, which require finer cross-modal granularity.
The bar chart on the right summarizes the overall improvement in mean accuracy, where the full model achieves the largest gain (+36.8) over the baseline, validating the complementarity of both stages.

\subsection{Qualitative Analysis}
Figure~\ref{fig:qualitative} presents qualitative comparisons across representative recommendation, recognition, and retrieval QA tasks.
In the recommendation example, existing models often misclassify a casual travel outfit as “social,” while OmniFashion accurately interprets the scene context and style.
In the recognition task, other models fail to identify the color of the shorts due to partial occlusion, whereas OmniFashion correctly infers “blue.”
In the retrieval cases, OmniFashion captures fine-grained cues such as the red rose embroidery or matches front–back views of the same outfit, where other models rely only on coarse color and texture similarity.
Notably, despite using only a 3B backbone, OmniFashion demonstrates stronger fashion understanding than larger 7B and even 32B counterparts.

\section{Discussion}
FashionX offers broad and well-structured supervision, yet its annotations inevitably carry the stylistic tendencies of the underlying VLM, reminding us that fully objective fashion labeling remains challenging.
The unified QA paradigm brings substantial coherence across recognition, reasoning, and retrieval, but richer linguistic understanding and long-context dialogue remain a challenge for future iterations.
Even with these considerations, OmniFashion’s performance across style reasoning, multi-image comparison, and retrieval tasks indicates encouraging potential for practical deployment in real-world fashion assistants, where both accuracy and interaction quality are essential.

\section{Conclusion}

In this work, we present FashionX and OmniFashion as two complementary contributions toward advancing generalist fashion intelligence.
FashionX offers large-scale, structured, and automatically generated annotations that address the incompleteness and inconsistency of existing datasets. Its unified, head-to-toe schema integrates global outfit semantics with fine-grained part attributes, providing a coherent foundation for multi-level fashion understanding.
Built upon this foundation, OmniFashion reformulates diverse fashion tasks,
including reasoning, attribute recognition, and cross-modal retrieval, into a unified fashion dialogue paradigm. This formulation connects detailed visual cues with holistic outfit interpretation and supports consistent learning across heterogeneous objectives. 
Surpassing all comparable open-source VLMs and remaining competitive with much stronger closed-source models across style reasoning, multi-image comparison, and fine-grained retrieval, OmniFashion shows that unified dialogue learning and structured fashion supervision can meaningfully advance VLMs toward practical generalist fashion intelligence.

\empty
\bibliographystyle{IEEEtran}
\bibliography{fashion}

\vfill

\end{document}